\documentclass[conference]{IEEEtran}
\IEEEoverridecommandlockouts

\usepackage{amsmath,amsfonts,graphicx}
\usepackage{caption}
\usepackage{subcaption}
\usepackage{hyperref}

\usepackage{tikz}
\usetikzlibrary{shapes,arrows,decorations.text, positioning}
\usepackage{etoolbox}
\usetikzlibrary{calc,shapes,arrows,decorations.text,positioning,patterns,matrix,chains,decorations.pathreplacing}
\usetikzlibrary{shapes.multipart}
\usepackage[]{xcolor}
\definecolor{violet_ensam}{RGB}{142,37,98}
\definecolor{orange_ensam}{RGB}{242,148,0}
\definecolor{gris_ensam}{RGB}{88,88,90}
\definecolor{gris_claire}{RGB}{220,220,220}

\title{Automatic Construction of Real-World Datasets for 3D Object Localization using Two Cameras}

\author{
\IEEEauthorblockN{Joris Gu\'erin, Olivier Gibaru, Eric Nyiri, St\'ephane Thiery and Jorge Palos}
\IEEEauthorblockA{LISPEN, Arts et M\'etiers ParisTech, Lille, France \\
contact: joris.guerin@ensam.eu}
}

\begin{document}
\maketitle


\begin{abstract}
Unlike classification, position labels cannot be assigned manually by humans. For this reason, generating supervision for precise object localization is a hard task. This paper details a method to create large datasets for 3D object localization, with real world images, using an industrial robot to generate position labels. By knowledge of the geometry of the robot, we are able to automatically synchronize the images of the two cameras and the object 3D position. We applied it to generate a screw-driver localization dataset with stereo images, using a KUKA LBR iiwa robot. This dataset could then be used to train a CNN regressor to learn end-to-end stereo object localization from a set of two standard uncalibrated cameras.

\end{abstract}



\section{Introduction}
In the context of autonomous manipulation, 3D object localization plays an essential role. Stereo vision is one of the most efficient methods for 3D reconstruction and thus is a very useful tool when dealing with robotic manipulation. Indeed, stereo vision has received many attention in research over the past decade in different subfields of autonomous manipulation (grasping \cite{grasping_stereo, grasping_stereo2}, contact-reach tasks \cite{manipulation_stereo}, and even playing soccer \cite{robocup_stereo}, ...).

However, classical methods of stereo vision are hard to design and require a lot of tuning. They can be inaccurate as they are composed of many steps, which are all potential  sources of errors (see section \ref{sec:stereo}). For this reason, we strongly believe that stereo localization would benefit from an end-to-end learning approach, which would simplify the design process and hopefully improve the system accuracy. Recently, several end-to-end approaches have been successful in achieving complex tasks, such as robot manipulation \cite{deepVisuo}, self-driving cars \cite{endtoend_driving}, speech recognition \cite{endtoend_speech}, obstacle avoidance \cite{endtoend_obstacle}... Even for end-to-end learning of manipulation tasks, which includes object localization, the vision part must be pretrained to locate objects if we want the reinforcement learning to be scalable to real life applications \cite{deepVisuo}. Hence, this work, which mainly focuses on end-to-end learning of 3D stereo object localization, also has great significance within the wider context of robotic autonomous learning of manipulation tasks.

To train a convolutional neural network (or any regression model) to learn stereo localization, we need to generate labelled data for localization (i.e., stereo images of the object to be located together with its position in space). Gathering labels for localization is a hard task. Position labels are hard to get as they cannot be written manually without spending precious time doing very precise measurements for each sample. In this paper, we introduce an approach for building such a dataset, by using an industrial robot to generate labelled data for 3D stereo localization. The main contribution of this paper is to describe a procedure to gather a lot of labelled stereo data for object localization automatically. This procedure is applied to generate a dataset for screw drivers 3D localization, composed of more than three thousand samples. Such a dataset\footnote{The dataset can be downloaded at \url{https://github.com/jorisguerin/SemanticViewSelection_dataset}.} also constitutes a progress as it will give a baseline for future research to test various regression models (CNN architectures) for stereo localization.


\section{Motivations for end-to-end stereo localization}
\label{sec:stereo}
    \subsection{Classical methods for 3D object localization using stereo vision}
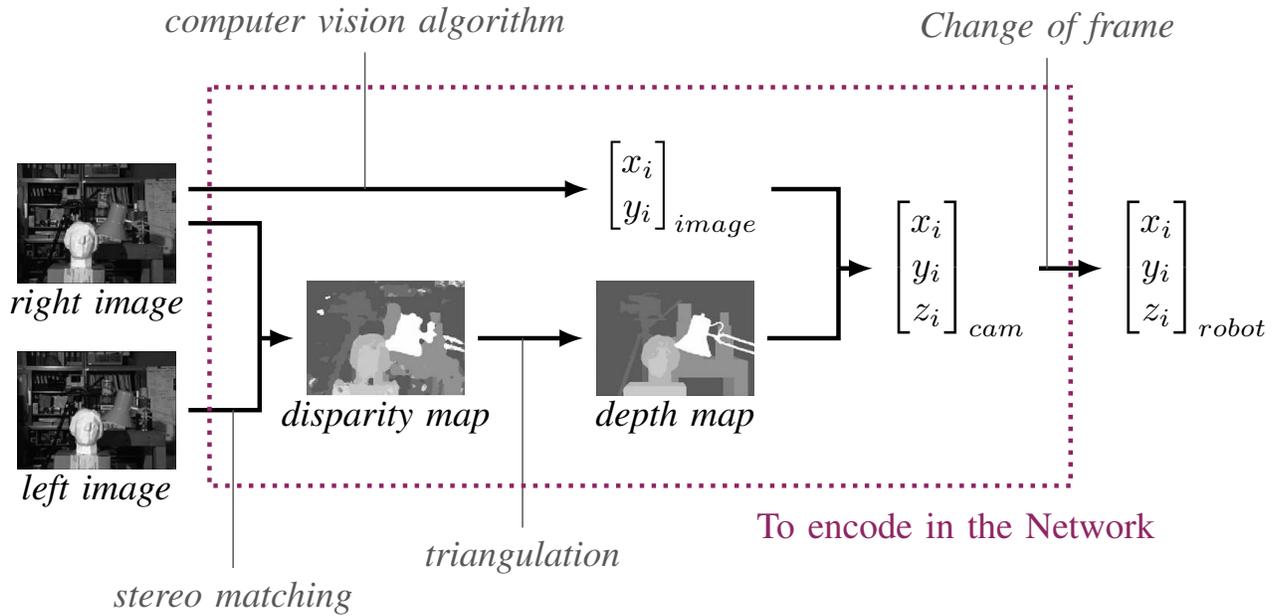
\begin{figure*}[!ht]
    \centering
    
    \newlength{\imwidth}
    \newlength{\labdist}
    \newlength{\hspacing}
    
    \setlength{\imwidth}{1.5cm}
    \setlength{\labdist}{-0.2cm}
    \setlength{\hspacing}{1cm}
    
    \tikzstyle{block} = [draw, rectangle, minimum width=0.1\textwidth, minimum height=0.1\textheight]
    
    \scalebox{1.4}{
    \begin{tikzpicture}[]
        \node[] (origin) at (0, 0) {};
        
        \node[above = 0.2cm of origin.center] (imageR) {\includegraphics[width = \imwidth]{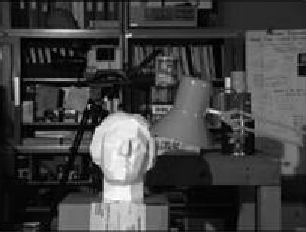}};
        \node[above = 0.2cm of imageR.east] (afterimR) {};
        \node[below = \labdist of imageR.south] (r_lab) {\textit{right image}};
        \node[below = 0.2cm of origin.center] (imageL) {\includegraphics[width = \imwidth]{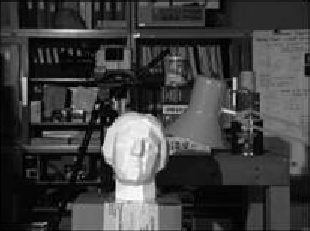}};
        \node[below = \labdist of imageL.south] (l_lab) {\textit{left image}};
        \node[right = 0.3\hspacing of imageL.east] (afterimL) {};
        
        \node[right = \hspacing of imageL.north east] (disparity) {\includegraphics[width = \imwidth]{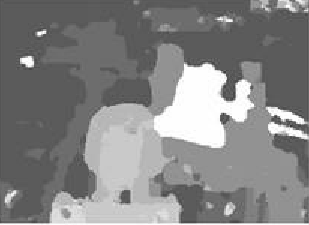}};
        \node[left = 0.2cm of disparity.west] (beforedisp) {};
        \node[below = \labdist of disparity.south] (disp_lab) {\textit{disparity map}};
        \node[right = 0.3\hspacing of disparity.east] (afterdisp) {};
        
        \node[right = \hspacing of disparity.east] (3dmap) {\includegraphics[width = \imwidth]{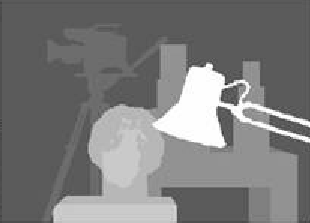}};
        \node[below = \labdist of 3dmap.south] (l_im) {\textit{depth map}};
        
        \node[right = 3.8\hspacing of afterimR.center] (2dpoints) {$\begin{bmatrix} x_i \\ y_i \end{bmatrix}_{image}$};
        \node[left = 2\hspacing of 2dpoints.west] (before2d) {};
        
        \node[right = \hspacing of 3dmap.north east] (3dptscam) {$\begin{bmatrix} x_i \\ y_i \\ z_i \end{bmatrix}_{cam}$};
        \node[left = 0.2cm of 3dptscam.west] (beforecam) {};
        
        \node[right = 0.6\hspacing of 3dptscam.east] (3dptsrob) {$\begin{bmatrix} x_i \\ y_i \\ z_i \end{bmatrix}_{robot}$};
        \node[left = 0.4cm of 3dptsrob.west] (beforerob) {};
        
        \node[gris_ensam, below = 1.5cm of afterimL.center] (stereomatch) {\textit{stereo matching}};
        \node[gris_ensam, below = 1.8cm of afterdisp.center] (triangulation) {\textit{triangulation}};
        \node[gris_ensam, above = 1.3cm of before2d.center] (cvalgo) {\textit{computer vision algorithm}};
        \node[gris_ensam, above = 2cm of beforerob.center] (framechange) {\textit{Change of frame}};
        
        \draw[very thick] (imageR.east) -| (beforedisp.center);
        \draw[very thick] (imageL.east) -| (beforedisp.center);
        \draw[very thick, ->, >=latex] (beforedisp.center) -- (disparity.west);
        
        \draw[very thick, ->, >=latex] (afterimR.center) -- (2dpoints.west);
        
        \draw[very thick, ->, >=latex] (disparity.east) -- (3dmap.west);
        
        \draw[very thick] (2dpoints.east) -| (beforecam.center);
        \draw[very thick] (3dmap.east) -| (beforecam.center);
        \draw[very thick, ->, >=latex] (beforecam.center) -- (3dptscam.west);
        
        \draw[very thick, ->, >=latex] (3dptscam.east) -- (3dptsrob.west);
        
        \draw[gris_ensam] (afterimL.center) -- (stereomatch.north);
        \draw[gris_ensam] (afterdisp.center) -- (triangulation.north);
        \draw[gris_ensam] (before2d.center) -- (cvalgo.south);
        \draw[gris_ensam] (beforerob.center) -- (framechange.south);
        
        \draw[violet_ensam, very thick,dotted] ($(imageR.north east)+(0.2,0.6)$) rectangle ($(3dptscam.south east)+(0.3,-1.3)$);
        
        \node[violet_ensam, below = 2.2cm of 3dptscam.center] (b) {To encode in the Network};
    \end{tikzpicture}}
    \caption{Typical pipeline for stereo vision object localization}
    \label{fig:stereo}
\end{figure*}

    The use of stereo vision has a long history in robotics manipulation \cite{grasping_stereo, manipulation_stereo}. However, currently, stereo localization consists in stacking different methods that can each be inaccurate. As we can see in figure \ref{fig:stereo}, to implement 3D object localization, it is needed to calibrate both cameras, compute accurate stereo matching, build a computer vision pipeline to identify pixels of interest, triangulate, and measure accurate transformation of frames to project the result in the frame of interest.
    
    More recently, researchers have started to use deep learning for stereo vision \cite{cnn_stereomatching1, cnn_stereomatching2, cnn_stereomatching3}. However, they mainly focus on stereo matching, which is only solving part of the problem. This trend can be seen by looking at the problem proposed with the most famous datasets for stereo vision \cite{kitti_stereomatching, middlebury}. To locate an object in space, one also needs to get the precise pixels in one of the image, which can cause errors, even with perfect stereo matching. Indeed, this problem is close to instance segmentation \cite{instance_segmentation} and is not easy. 
    In general, end-to-end learning approaches tend to be better than stacking subsystems as it can correct internal errors. For this reason we introduce such a framework for stereo localization in the next section. This paper is a first step towards implementing it as it proposes a generic way to produce real-world datasets for stereo object localization using a robot to generate position labels. 
    
    \subsection{End-to-end pipeline for stereo localization}
    
    An end-to-end pipeline for stereo object localization would consist in mapping pixels from two images to 3D points in space representing an object pose, using a regression function approximator. Such a pipeline could then be used for grasping, or any robotic manipulation task.
    
    In this paper, we propose an approach to build datasets for stereo localization learning. Showing the feasibility of gathering large datasets, is a first step towards investigating the feasibility of end-to-end object stereo localization. The purple dotted line in figure \ref{fig:stereo} illustrates what the network is supposed to encode (the relative camera calibration could be added). After seeing the amazing results produced by CNNs when dealing with images, it seems that only the difficulty to gather enough labelled data prevents them to carry-out end-to-end stereo localization. The idea we implement in this paper consists in defining a method, using a precise and accurate industrial robot, to generate labelled data for object localization. We also apply it to the get data for screw-drivers localization. Section \ref{sec:dataset} contains more details about the dataset .
	

\section{Dataset generation}
\label{sec:dataset}
    When building a dataset for supervised learning, it is crucial to make sure that our input data are sufficient to know everything about our outputs. For the case of stereo object localization, all that is required is to have two cameras with fixed focal length and relative positions. We also need to keep the position of the two cameras fixed with respect to the robot and to make sure that both cameras are oriented such that they share partially common fields of view (the object must be seen by both cameras).
    
    \subsection{Generate diversity to avoid overfitting}
    
    In order to avoid overfitting, the dataset should have as much diversity as possible. To do so, we try to change the following parameters:
    \begin{itemize}
        \item lighting conditions, by gathering the data in a shop-floor with unmastered lighting.
        \item background, by placing different clothes/objects on the table at which cameras are looking
    \end{itemize}
    We also add images where we place distractor objects (other tools) in the background so that the CNN would learn to locate screw drivers and not any object. The dataset contains several different screwdrivers to help him discover the concept of what makes a screw driver.
    
\begin{figure*}[t]
        \centering
        \includegraphics[width = 0.22\textwidth]{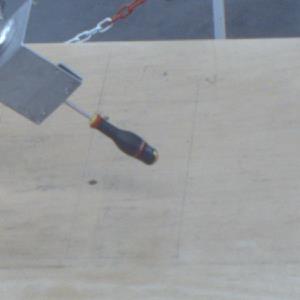} \hspace{1pt}
        \includegraphics[width = 0.22\textwidth]{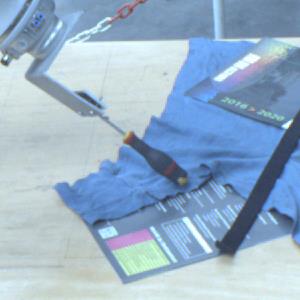} \hspace{1pt}
        \includegraphics[width = 0.22\textwidth]{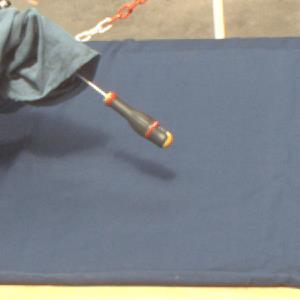} \hspace{1pt}
        \includegraphics[width = 0.22\textwidth]{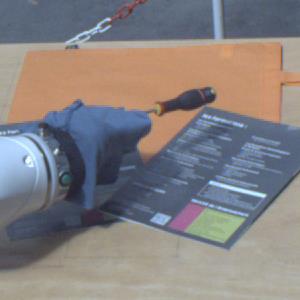}\\ \vspace{5pt}
        \includegraphics[width = 0.22\textwidth]{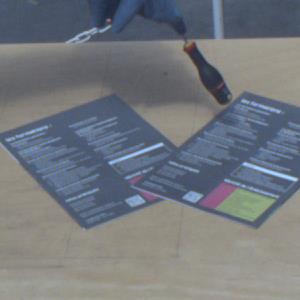} \hspace{1pt}
        \includegraphics[width = 0.22\textwidth]{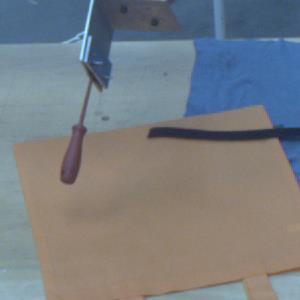} \hspace{1pt}
        \includegraphics[width = 0.22\textwidth]{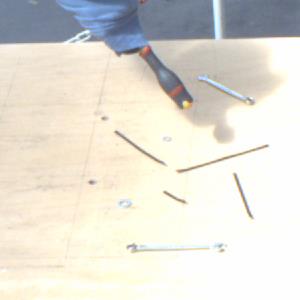} \hspace{1pt}
        \includegraphics[width = 0.22\textwidth]{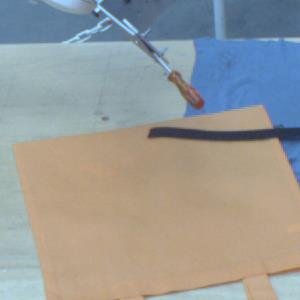}
        \caption{Representative subset of the images from the screw driver localization dataset.}
        \label{fig:images}
    \end{figure*}

    Finally, we make sure that the random configurations of the robot explore the full range of positions and orientations allowed within the common view range of the cameras. Figure \ref{fig:images} shows a representative subset of the images present in the dataset. Note that each image shown has a corresponding image from the other camera.

    \subsection{Avoid learning the wrong thing by removing the robot}

    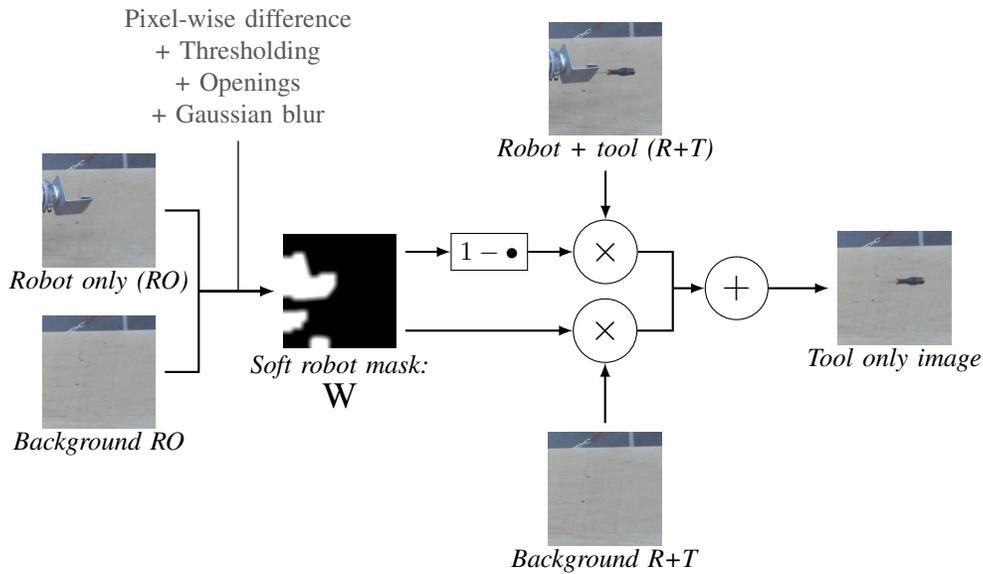
\begin{figure*}[!ht]
    \centering
    
    \setlength{\imwidth}{1.5cm}
    \setlength{\labdist}{-0.2cm}
    \setlength{\hspacing}{1.5cm}
    
    \tikzstyle{block} = [draw, rectangle]
    
    \tikzset{%
  sum/.style      = {draw, circle, node distance = 2cm}, 
}
\newcommand{\suma}{\Large$+$}
\newcommand{\multa}{\Large$\times$}

\scalebox{1}{
    \begin{tikzpicture}[]
        \node[] (origin) at (0, 0) {};
        
        \node[above = 0.2cm of origin.center] (noObj) {\includegraphics[width = \imwidth]{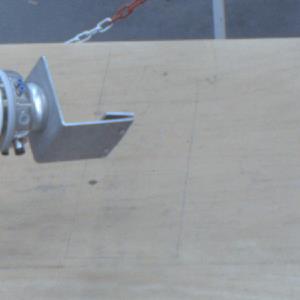}};
        \node[below = 0.2cm of origin.center] (bg_noObj) {\includegraphics[width = \imwidth]{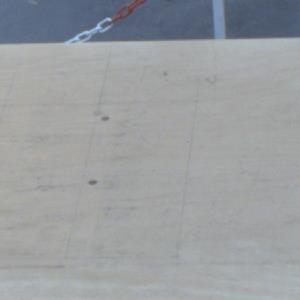}};
        \node[below = \labdist of noObj.south] (noObj_lab) {\textit{Robot only (RO)}};
        \node[below = \labdist of bg_noObj.south] (bg_noObj_lab) {\textit{Background RO}};
        
        \node[right = 1.2cm of origin.center] (inter1) {};
        \node[right = 0.4cm of inter1.center] (inter12) {};
        
        \node[gris_ensam, above = 2cm of inter12.center] (arrow_lab) {\begin{tabular}{c} Pixel-wise difference \\ + Thresholding \\ + Openings \\ + Gaussian blur\end{tabular}};
        
        \draw[thick, gris_ensam] (arrow_lab.south) -- (inter12.center);
        \draw[thick] (noObj.east) -| (inter1.center);
        \draw[thick] (bg_noObj.east) -| (inter1.center);
        
        \node[right = 1cm of inter1.center] (robotMask) {\includegraphics[width = \imwidth]{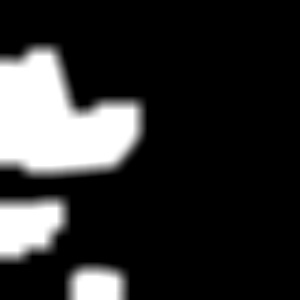}};
        \node[below = \labdist of robotMask.south] (robotmask_lab) {\begin{tabular}{c} \textit{Soft robot mask:} \\ \Large{W} \end{tabular}};
        
        \draw[thick, ->, >=latex] (inter1.center) -- (robotMask.west);
        
        \node[right = 1.8cm of noObj.east] (inter2) {};
        
        \node[above = 0.4cm of robotMask.east] (inter5) {};
        \node[block, right = 0.6cm of inter5.center] (oneminus) {$1-\bullet$};
        \node[sum, right = 0.6cm of oneminus.east] (mult1) {\multa};
        
        \draw[thick, ->, >=latex] (inter1.center) -- (robotMask.west);
        \draw[thick, ->, >=latex] (inter1.center) -- (robotMask.west);
        
        \node[below = 0.4cm of robotMask.east] (inter6) {};
        \node[sum, right = 2.1cm of inter6.east] (mult2) {\multa};
    
        \draw[thick, ->, >=latex] (inter5.center) -- (oneminus.west);
        \draw[thick, ->, >=latex] (oneminus.east) -- (mult1.west);
        \draw[thick, ->, >=latex] (inter6.center) -- (mult2.west);
        
        \node[above = 1cm of mult1.north] (obj) {\includegraphics[width = \imwidth]{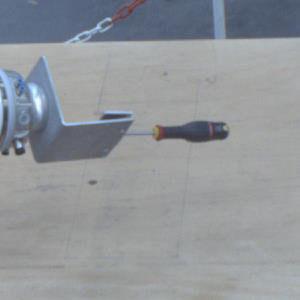}};
        \node[below = 0.8cm of mult2.south] (bg_obj) {\includegraphics[width = \imwidth]{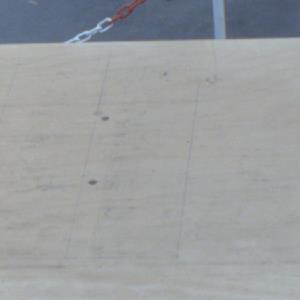}};
        \node[below = \labdist of obj.south] (obj_lab) {\textit{Robot + tool (R+T)}};
        \node[below = \labdist of bg_obj.south] (bg_obj_lab) {\textit{Background R+T}};
        
        \draw[thick, ->, >=latex] (obj_lab.south) -- (mult1.north);
        \draw[thick, ->, >=latex] (bg_obj.north) -- (mult2.south);
        
        \node[above = 0.01cm of mult2.north] (inter3){};
        \node[sum, right = 1.2cm of inter3.east] (sum) {\suma};
        
        \node[left = 0.3cm of sum.west] (inter4){};
        \draw[thick] (mult1.east) -| (inter4.center);
        \draw[thick] (mult2.east) -| (inter4.center);
        \draw[thick, ->, >=latex] (inter4.center) -- (sum.west);
        
        \node[right = 0.8cm of sum.east] (res) {\includegraphics[width = \imwidth]{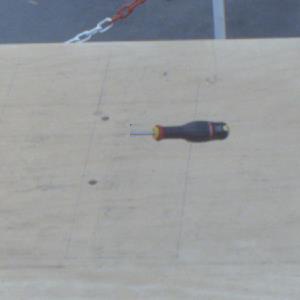}};
        \node[below = \labdist of res.south] (res_lab) {\textit{Tool only image}};
        
        \draw[thick, ->, >=latex] (sum.east) -- (res.west);
    \end{tikzpicture}}
    
    \caption{Computer vision pipeline for removing the robot from four images}
    \label{fig:cvpipeline}
\end{figure*}
    The idea of using a robot to generate supervision opens a broad range of possibilities for object localization. However, there can be a drawback, the robot (or at least the tool holder) will appear on every image and we can fear that the regressor learns to locate the robot and just applies some kind of shift to find the screwdriver. To avoid such problem, we need to remove the robot from some of the images. We proceed in two different ways:
    \begin{itemize}
        \item Physically, we hide the robot by wrapping some cloth around it (figure \ref{fig:images})
        
        \item Computationally, using a computer vision (CV) pipeline to remove the robot from images. To do so, we increase the dataset and take four images per sample (robot + tool, robot alone, and the background corresponding to the two situations). The full pipeline is described in Figure \ref{fig:cvpipeline}.
    \end{itemize}
    
    The underlying idea behind the robot removal CV algorithm is to use the robot only image and the background to compute a mask of the robot and replace the corresponding pixels by background pixels. A similar approach has been proposed by \cite{levine_grasping}. The necessity to get two background images comes from the changes in lighting conditions. This algorithm is then fine tuned to remove only the good pixels and get smooth edges.
    
    To ensure that all the images in our dataset do not contain any systemic patterns coming from our robot removal algorithm, it is also important to keep original images, i.e. with the robot. 
    
    \subsection{Technical details on data generation}
    
    For tool calibration, we assume that the screw driver is orthogonal to the tool holder and we use the very accurate torque sensors of the Kuka LBR iiwa robot to detect contact reaches with a plaque in several directions. In this way, we can get all the shifts and compute the transformation of frames.
    
    When we generate data, we want to make sure that the object stands within a certain cuboid (the biggest inside the common views), that the robot is not hiding the object in the image, and obviously, that the robot is not colliding with itself or the environment.
    
    We also check, for the cases where we take several images, that the robot is not behind the screw driver so that there is no mask superposition and we do not remove part of the tool with our computer vision algorithm.


\section{Conclusion}
\label{sec:conclusion}
In this paper, we described a procedure to build a dataset for 3D object localization using stereo vision. By using an industrial robot, we have shown that it is possible to generate data with accurate position labels. We propose an approach to generate data with enough variability, to avoid overfitting, as well as a method to remove the robot from some of the images to prevent our regressor from learning the wrong thing. We also applied it to build a dataset for screw driver localization.

Obviously, the next steps for this work will consist in exploiting this dataset and design a CNN architecture to encode the 3D stereo localization pipeline. Another perspective is to apply this process as a preprocessing step for manipulation reinforcement learning, likewise in \cite{deepVisuo}. By doing so, we can expect to get better results with a policy using stereo images instead of a single camera.

\section*{
ACKNOWLEDGMENTS}
This work was supported by the European Union's 2020 research and innovation program under grant agreement No.688807, project ColRobot (collaborative robotics for assembly and kitting in smart manufacturing).

\bibliographystyle{IEEEtran}
\bibliography{IEEEabrv,biblio.bib}

\begin{thebibliography}{10}
\providecommand{\url}[1]{#1}
\csname url@samestyle\endcsname
\providecommand{\newblock}{\relax}
\providecommand{\bibinfo}[2]{#2}
\providecommand{\BIBentrySTDinterwordspacing}{\spaceskip=0pt\relax}
\providecommand{\BIBentryALTinterwordstretchfactor}{4}
\providecommand{\BIBentryALTinterwordspacing}{\spaceskip=\fontdimen2\font plus
\BIBentryALTinterwordstretchfactor\fontdimen3\font minus
  \fontdimen4\font\relax}
\providecommand{\BIBforeignlanguage}[2]{{%
\expandafter\ifx\csname l@#1\endcsname\relax
\typeout{** WARNING: IEEEtran.bst: No hyphenation pattern has been}%
\typeout{** loaded for the language `#1'. Using the pattern for}%
\typeout{** the default language instead.}%
\else
\language=\csname l@#1\endcsname
\fi
#2}}
\providecommand{\BIBdecl}{\relax}
\BIBdecl

\bibitem{grasping_stereo}
P.~Azad, T.~Asfour, and R.~Dillmann, ``Stereo-based 6d object localization for
  grasping with humanoid robot systems,'' in \emph{Intelligent Robots and
  Systems, 2007. IROS 2007. IEEE/RSJ International Conference on}.\hskip 1em
  plus 0.5em minus 0.4em\relax IEEE, 2007, pp. 919--924.

\bibitem{grasping_stereo2}
A.~Morales, T.~Asfour, P.~Azad, S.~Knoop, and R.~Dillmann, ``Integrated grasp
  planning and visual object localization for a humanoid robot with
  five-fingered hands,'' in \emph{Intelligent Robots and Systems, 2006 IEEE/RSJ
  International Conference on}.\hskip 1em plus 0.5em minus 0.4em\relax IEEE,
  2006, pp. 5663--5668.

\bibitem{manipulation_stereo}
N.~Hudson, T.~Howard, J.~Ma, A.~Jain, M.~Bajracharya, S.~Myint, C.~Kuo,
  L.~Matthies, P.~Backes, P.~Hebert \emph{et~al.}, ``End-to-end dexterous
  manipulation with deliberate interactive estimation,'' in \emph{Robotics and
  Automation (ICRA), 2012 IEEE International Conference on}.\hskip 1em plus
  0.5em minus 0.4em\relax IEEE, 2012, pp. 2371--2378.

\bibitem{robocup_stereo}
U.-P. K{\"a}ppeler, M.~H{\"o}ferlin, and P.~Levi, ``3d object localization via
  stereo vision using an omnidirectional and a perspective camera,'' in
  \emph{Proceedings of the 2nd Workshop on Omnidirectional Robot Vision,
  Anchorage, Alaska}, 2010, pp. 7--12.

\bibitem{deepVisuo}
S.~Levine, C.~Finn, T.~Darrell, and P.~Abbeel, ``End-to-end training of deep
  visuomotor policies,'' \emph{arXiv preprint arXiv:1504.00702}, 2015.

\bibitem{endtoend_driving}
M.~Bojarski, D.~Del~Testa, D.~Dworakowski, B.~Firner, B.~Flepp, P.~Goyal, L.~D.
  Jackel, M.~Monfort, U.~Muller, J.~Zhang \emph{et~al.}, ``End to end learning
  for self-driving cars,'' \emph{arXiv preprint arXiv:1604.07316}, 2016.

\bibitem{endtoend_speech}
D.~Amodei, S.~Ananthanarayanan, R.~Anubhai, J.~Bai, E.~Battenberg, C.~Case,
  J.~Casper, B.~Catanzaro, Q.~Cheng, G.~Chen \emph{et~al.}, ``Deep speech 2:
  End-to-end speech recognition in english and mandarin,'' in
  \emph{International Conference on Machine Learning}, 2016, pp. 173--182.

\bibitem{endtoend_obstacle}
U.~Muller, J.~Ben, E.~Cosatto, B.~Flepp, and Y.~L. Cun, ``Off-road obstacle
  avoidance through end-to-end learning,'' in \emph{Advances in neural
  information processing systems}, 2006, pp. 739--746.

\bibitem{cnn_stereomatching1}
W.~Luo, A.~G. Schwing, and R.~Urtasun, ``Efficient deep learning for stereo
  matching,'' in \emph{Proceedings of the IEEE Conference on Computer Vision
  and Pattern Recognition}, 2016, pp. 5695--5703.

\bibitem{cnn_stereomatching2}
P.~Kn{\"o}belreiter, C.~Reinbacher, A.~Shekhovtsov, and T.~Pock, ``End-to-end
  training of hybrid cnn-crf models for stereo,'' \emph{arXiv preprint
  arXiv:1611.10229}, 2016.

\bibitem{cnn_stereomatching3}
J.~Zbontar and Y.~LeCun, ``Stereo matching by training a convolutional neural
  network to compare image patches,'' \emph{Journal of Machine Learning
  Research}, vol.~17, no. 1-32, p.~2, 2016.

\bibitem{kitti_stereomatching}
A.~Geiger, P.~Lenz, and R.~Urtasun, ``Are we ready for autonomous driving? the
  kitti vision benchmark suite,'' in \emph{Conference on Computer Vision and
  Pattern Recognition (CVPR)}, 2012.

\bibitem{middlebury}
\BIBentryALTinterwordspacing
D.~Scharstein, H.~Hirschmüller, Y.~Kitajima, G.~Krathwohl, N.~Nesic, X.~Wang,
  and P.~Westling, ``High-resolution stereo datasets with subpixel-accurate
  ground truth.'' in \emph{GCPR}, ser. Lecture Notes in Computer Science,
  X.~Jiang, J.~Hornegger, and R.~Koch, Eds., vol. 8753.\hskip 1em plus 0.5em
  minus 0.4em\relax Springer, 2014, pp. 31--42. [Online]. Available:
  \url{http://dblp.uni-trier.de/db/conf/dagm/gcpr2014.html#ScharsteinHKKNWW14}
\BIBentrySTDinterwordspacing

\bibitem{instance_segmentation}
B.~Romera-Paredes and P.~H.~S. Torr, ``Recurrent instance segmentation,'' in
  \emph{European Conference on Computer Vision}.\hskip 1em plus 0.5em minus
  0.4em\relax Springer, 2016, pp. 312--329.

\bibitem{levine_grasping}
S.~Levine, P.~Pastor, A.~Krizhevsky, J.~Ibarz, and D.~Quillen, ``Learning
  hand-eye coordination for robotic grasping with deep learning and large-scale
  data collection,'' \emph{The International Journal of Robotics Research}, p.
  0278364917710318, 2016.

\end{thebibliography}

\end{document}